\documentclass{article} 
\usepackage{iclr2023_conference,times}

\usepackage{amsmath,amsfonts,bm}

\def\eqref#1{equation~\ref{#1}}

\def\1{\bm{1}}

\DeclareMathAlphabet{\mathsfit}{\encodingdefault}{\sfdefault}{m}{sl}
\SetMathAlphabet{\mathsfit}{bold}{\encodingdefault}{\sfdefault}{bx}{n}

\newcommand{\R}{\mathbb{R}}

\usepackage{hyperref}
\usepackage{url}

\usepackage{booktabs}
\usepackage{graphicx}
\usepackage{subfigure}
\usepackage{graphicx}
\usepackage{wrapfig}
\usepackage{multirow}

\title{Task-customized Masked AutoEncoder via Mixture of Cluster-conditional Experts}

\author{
    Zhili Liu\textsuperscript{\rm 1,2*},
    Kai Chen\textsuperscript{\rm 1*},
    Jianhua Han\textsuperscript{\rm 2},
    Lanqing Hong\textsuperscript{\rm 2},
    Hang Xu\textsuperscript{\rm 2}, \\
    \textbf{    
    Zhenguo Li\textsuperscript{\rm 2},
    James T. Kwok\textsuperscript{\rm 1}}\\
\textsuperscript{\rm 1}  Department of Computer Science and Engineering, Hong Kong University of Science and Technology\\
\textsuperscript{\rm 2} Huawei Noah’s Ark Lab\\
\texttt{\{zhili.liu,  kai.chen\}@connect.ust.hk,}\\
\texttt{\{hanjianhua4, honglanqing, xu.hang, li.zhenguo\}@huawei.com} \\
\texttt{jamesk@cse.ust.hk}\\
}

\iclrfinalcopy 
\begin{document}

\maketitle
\vspace{-4mm}

\def\thefootnote{*}\footnotetext{Equal contribution.}\def\thefootnote{\arabic{footnote}}


\begin{abstract}
Masked Autoencoder~(MAE) is a prevailing self-supervised learning method that achieves promising results in model pre-training.
However, when the various downstream tasks have data distributions different from the pre-training data, the semantically irrelevant pre-training information might result in negative transfer, impeding MAE's scalability.
To address this issue, we propose a novel MAE-based pre-training paradigm, Mixture of Cluster-conditional Experts (MoCE), which can be trained once but provides customized pre-training models for diverse downstream tasks. 
Different from the mixture of experts (MoE), our MoCE trains each expert only with semantically relevant images by using cluster-conditional gates.
Thus, each downstream task can be allocated to its customized model pre-trained with data most similar to the downstream data.
Experiments
on a collection of 11 downstream tasks
show that MoCE 
outperforms 
the vanilla MAE 
by 2.45\% on average. It also obtains
new state-of-the-art 
self-supervised 
learning results on detection and segmentation.
\end{abstract}


\section{Introduction}
Self-supervised learning (SSL), which learns effective transferable representations without human annotations, has become a prevailing model pre-training paradigm~\citep{he2020momentum,chen2021multisiam,bao2021beit}.
Currently, the most prevalent SSL method is the Masked Autoencoder (MAE)~\citep{he2022masked}, which constructs supervision signals from raw image data by masking random input patches and then reconstructing the missing
pixels.
This simple strategy has proved efficient in the training of large-scale models. For example,
ViT~\citep{dosovitskiy2020image} shows impressive performance on popular benchmarks such as the ImageNet
~\footnote{We refer to ImageNet-1K as ImageNet if not specified in this paper.}~\citep{deng2009imagenet}.
However, does MAE really scale well for various downstream tasks~\citep{deng2009imagenet,lin2014microsoft,ADE20K,han2021soda10m,li2022coda}?

\begin{figure*}[tbp] 
	\centering
	{\subfigure[Negative transfer phenomenon on MAE. ]{\includegraphics[width=0.55\columnwidth]{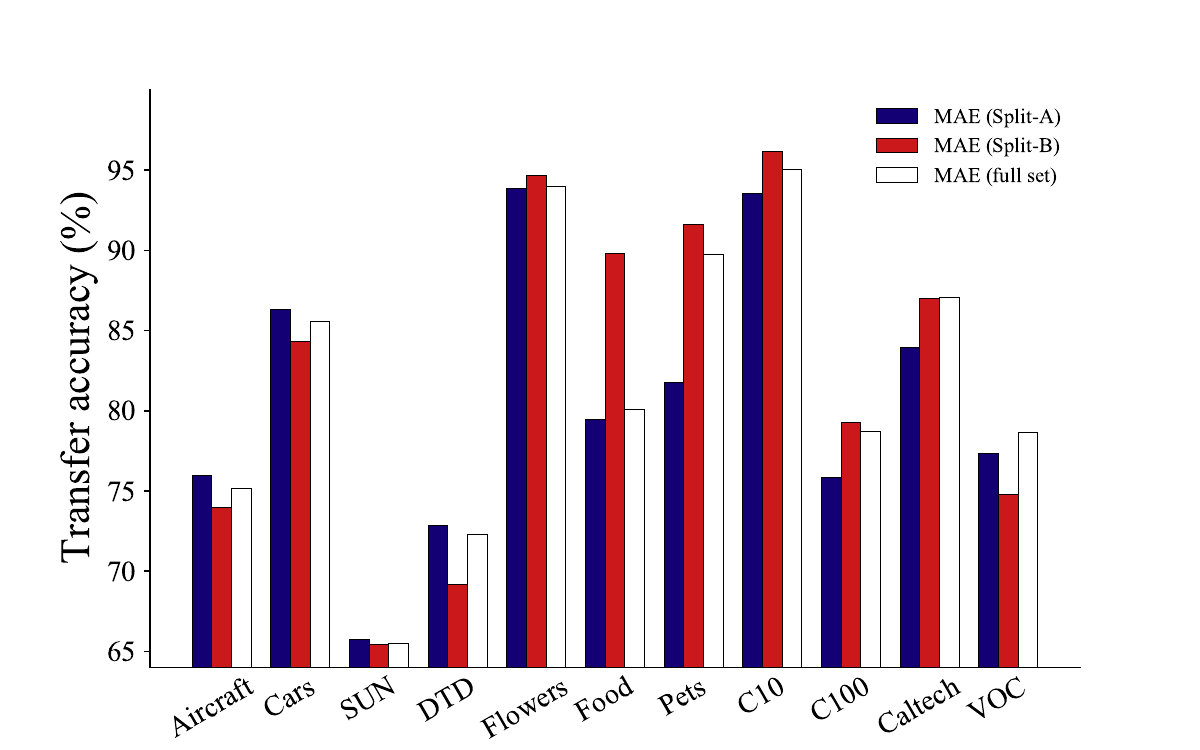}  \label{fig:neg_tras}}}\hspace{3mm}
	{\subfigure[Problem with TokenMoE. ]{\includegraphics[width=0.39\columnwidth ]{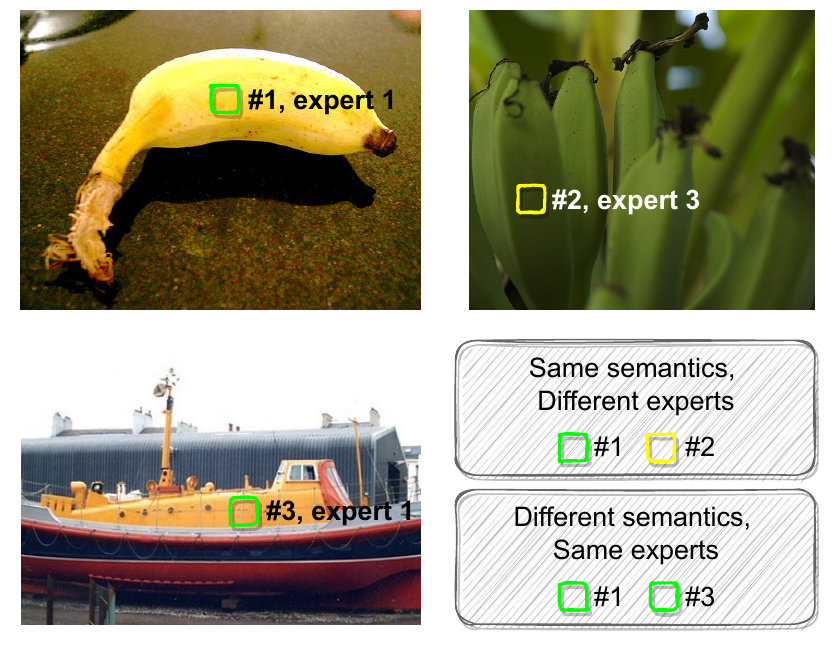}  \label{fig:disadv_tokenmoe}}}
\caption{(a) Transfer performance of MAEs pre-trained on 
Split-A~(blue), Split-B~(red) and full ImageNet data~(white).
        Only two of the eleven downstream tasks benefit from using the full
		  ImageNet data for pre-training
(more details in Section~\ref{sec:negative_transfer}). 
	(b) 
	 TokenMoE 
	 uses pixel RGB values as reconstruction targets. Thus, tokens with
	 similar pixel values tend to be routed to the same expert, leading to
    two types of mistakes: (i) same semantics but routed to different experts, 
	 (ii) different semantics but routed to the same expert.
	}
	\label{fig:negative_transfer}
\end{figure*}

Preliminary studies (in Section~\ref{sec:negative_transfer})
show that the MAE indeed suffers from {\it negative transfer}~\citep{liu2022task} when transferring to downstream tasks with very different semantics.
Figure~\ref{fig:neg_tras} shows that on 9 of 11 downstream tasks, an MAE pre-trained on the full ImageNet data is outperformed by the one that is pre-trained on only the semantically relevant data subsets.
Hence, using pre-training data that are semantically irrelevant can hurt transfer performance.

The above observation motivates the need for task-customized pre-training. 
A promising model for this is the
Mixture of Experts (MoE)~\citep{shazeeroutrageously, riquelme2021scaling}, which uses a multi-expert architecture to provide customized models for different input tokens. 
However, unlike supervised pre-training, 
self-supervised learning lacks semantic labels, and thus the experts differ more on low-level information than semantics, referring to Figure~\ref{fig:disadv_tokenmoe}. 
Experiments in Section~\ref{sec:analysis} show that a naive adoption of MoE to the MAE has inferior performance. Since various downstream tasks contain different semantics, semantic-related experts may be preferred.

In this paper, we propose the Mixture of Cluster-conditional Expert (MoCE), a novel paradigm to achieve task-customized self-supervised pre-training by data clustering and explicitly training each expert with images of similar semantics.
The MoCE procedure has three stages. First, we cluster the whole
dataset by using a pre-trained, dense MAE model.
We then construct the MoCE with a multi-expert structure. 
Each expert
is trained 
using clusters selected by routing tokens based on {\it cluster embedding}
(instead of {\it token embedding}). 
To stabilize training and
enhance confidence of the gate results,
a regularization loss is proposed.
Finally, with the arrival of a downstream task, we propose a search procedure to
select the
closest cluster.
Empirically, the proposed MoCE shows superior performance 
over MAE
on a collection of 11 downstream tasks. Besides, 
one can use only a
MoCE sub-model
on deployment, 
thus
saving inference time and model capacity.

To summarize, our main contributions are:
\begin{enumerate}
\item We systematically analyze the negative transfer phenomenon of MAE, and show
that naively adopting the MoE to MAE cannot improve transfer performance of downstream tasks.
\item We propose the MoCE, which trains each expert with semantics-aware clusters
so that
similar clusters 
can be routed to
the same expert.  
\item We demonstrate effectiveness of the proposed MoCE on a collection of 11
downstream tasks, and achieve up to 2.45\% performance improvement in Top-1 accuracy. State-of-the-art 
self-supervised 
results 
are also achieved on the detection and segmentation tasks. To the best of our
knowledge, this is the first work that 
achieves state-of-the-art transfer performance
by training vision MoE models with ImageNet under the SSL setting.
\end{enumerate}

\section{Related work}

\paragraph{Self-supervised Learning.}
Previous works mainly focus on the design of pretext tasks with image transformations~\citep{doersch2015unsupervised,gidaris2018unsupervised}, inpainting~\citep{pathak2016context}, colorization~\citep{zhang2016colorful}, contrastive learning~\citep{chen2020simple,he2020momentum,byol20,caron2020unsupervised,radford2021learning,yaofilip}, and for specific downstream tasks~\citep{wang2020dense,xie2020propagate,xie2021detco,chen2021multisiam,yaodetclip}.
Motivated by the design of BERT~\citep{devlin2018bert}, masked image modeling (MIM) is recently proposed to learn by reconstructing masked images.
BEiT~\citep{bao2021beit} is the pioneering work that predicts visual tokens generated by a pre-trained tokenizor~\citep{2021Learning}.
SimMIM~\citep{SimMIM} simplifies the framework by directly utilizing the pixel RGB
values as reconstruction targets. MAE~\citep{he2022masked} proposes an asymmetric encoder-decoder architecture for better training efficiency.
MixedAE~\citep{chen2023mixed} further explores image mixing for object-aware pre-training.
In this paper, we will
focus 
on the MAE 
due to its effectiveness and efficiency.

While 
self-supervised learning methods
have achieved improved transfer performance, 
most of them only provide a unified representation to various downstream tasks.
This may suffer from negative transfer as demonstrated in Section~\ref{sec:negative_transfer}.
The work most relevant to ours is SDR~\citep{liu2022task}, which trains 256 subnets with 256 disjoint ImageNet subsets simultaneously. 
However, this paper differs from 
SDR
in three ways: (i) 
the mapping from
subsets 
to subnets 
in SDR
is randomly selected and fixed during pre-training, while MoCE achieves self-adaptive mapping with cluster-conditional gates; 
(ii) Progressive training is required in SDR, while MoCE enjoys one-time end-to-end training;
(iii) During the transfer process, SDR uses
brute force
to select the best sub-model, while MoCE reuses the clustering module to achieve more efficient selection.

\paragraph{Mixture of Experts.}
The mixture of experts (MoE) has a long
history~\citep{jacobs1991adaptive,jordan1994hierarchical,shazeeroutrageously}.
Recently, it is considered as an effective tool for 
model scale-up in
natural language
processing~\citep{lepikhin2020gshard,fedus2021switch,yang2021m6,lewis2021base}.
With the growing interest of the Vision Transformer
\citep{dosovitskiy2020image,liu2021swin,wang2021pyramid,xie2021segformer}, MoE for
vision~\citep{riquelme2021scaling,wu2022residual} is also explored recently.
However, there is still no self-supervised MoE model that can be trained on 
medium-sized
datasets such as the ImageNet-1k.

\cite{kudugunta2021beyond,ma2018modeling} 
regard the MoE as a multi-task learning model, and use it for multi-language
translation and recommendation systems, respectively. In this paper, we show that
for 
self-supervised learning on images, an additional clustering
component is crucial in the learning of a highly performant MoE model.
Moreover, 
while the downstream tasks should follow the pre-training task
in \citep{kudugunta2021beyond,ma2018modeling},
the MoCE can be used with
any downstream task due to
its unsupervised pre-training.
\cite{puigcerver2020scalable} shares a similar setting with us, but their
model is pre-trained in a supervised learning manner. Moreover, their mapping
between experts and data is pre-defined and fixed during training, while that for
the MoCE is learned dynamically and achieves better performance.

\paragraph{Multi-Task Learning} aims to learn a model that is appropriate for multiple tasks. 
Hard-parameter sharing, which uses a shared backbone with multi-heads for the different tasks, has been 
shown to be effective on time series,
language and graph data
\citep{liu2019multi,hu2019strategies,mcdermott2021comprehensive}.
\cite{gao2021network} claims that the network design may further benefit from the use of 
task relationships, and trains masks for different tasks. However,
they require the task information be available during model training, which is not
possible for
downstream tasks in SSL pre-training.

\section{Proposed Method}
\label{sec:method}

In this section, we first empirically demonstrate the negative transfer phenomenon in MAE 
(Section~\ref{sec:negative_transfer}). 
We then discuss the limitations of adopting TokenMoE~\citep{riquelme2021scaling}
with MAE (Section~\ref{sec:tokenmoe}), and propose the Mixture of
Cluster-conditional Experts (MoCE), a novel paradigm achieving customized
pre-training for various downstream tasks (Section~\ref{sec:moce}).


\subsection{Negative Transfer in Masked AutoEncoder}
\label{sec:negative_transfer}


\begin{figure}[t!]
    \centering
    \includegraphics[width=1.0\linewidth]{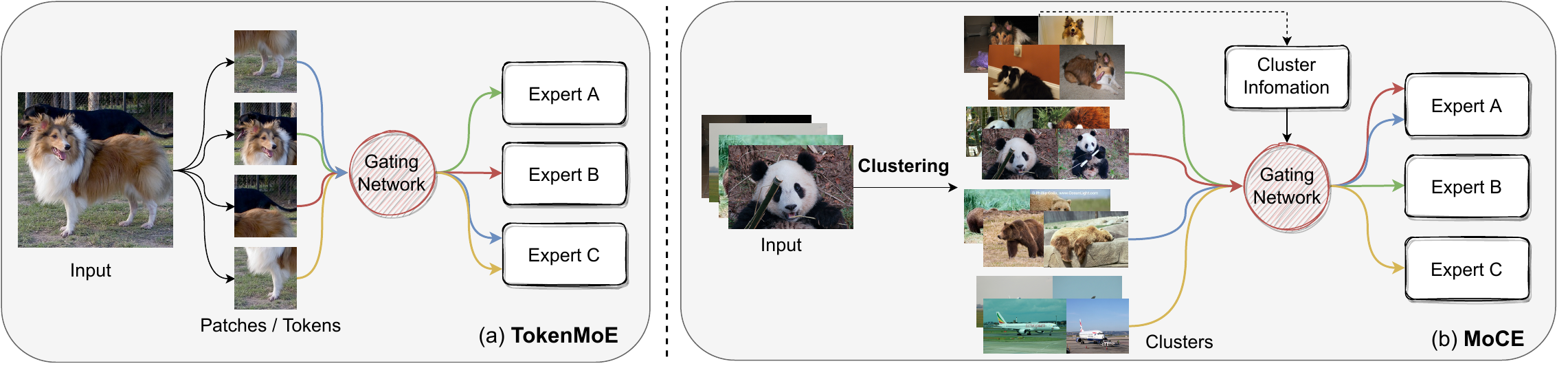}
    \caption{Model design comparison between (a) TokenMoE~\citep{riquelme2021scaling} and (b) MoCE. 
    Both methods utilize the multi-expert architecture with the main difference about the input of the gating network. MoCE adopts the corresponding cluster embedding of the current token as in Eqn.~\ref{equ:moce}, instead of the token embedding in Eqn.~\ref{equ:tokenmoe}. 
    Therefore, each expert can be trained by semantically similar images to alleviate the negative transfer phenomenon. 
    }
    \label{fig:arch_comparison}
\end{figure}


In this section, we evaluate the transfer performance of MAE models pre-trained
with data of different semantics on various downstream tasks.
As in \citep{huh2016makes,liu2022task},
we first split the ImageNet data into two disjoint subsets, Split-A and
Split-B, based on the labels' semantic dissimilarities in the WordNet tree~\citep{miller1998wordnet}.
Split-A mainly contains inanimate objects (such as cars and airplanes), while
Split-B primarily involves organisms (such as plants and animals).
We then pre-train MAEs on Split-A, Split-B and the full ImageNet without data annotation, and evaluate the three resulting models on 11 downstream tasks.
See more implementation details in Section~\ref{sec:setup}.
 
As shown in Figure~\ref{fig:neg_tras}, the MAE pre-trained with Split-A performs
best on {\it Aircraft}, {\it Cars}, {\it SUN397} and {\it DTD}, while  the MAE
pre-trained with Split-B
performs best on {\it Flowers}, {\it Food}, {\it Pets}, {\it Cifar-10} and {\it
Cifar-100}.
Only two of the eleven tasks 
({\it Caltech} and {\it VOC})
benefit from using the full ImageNet data. This
suggests that for tasks whose semantics are close to inanimate objects,
adding pre-training data from Split-B is not useful,
and vice versa for tasks whose semantics are close to organisms.
To conclude, the introduction of semantically irrelevant pre-training data may impede transfer performance for downstream tasks.
This negative transfer phenomenon motivates us to develop an efficient and automatic paradigm for task-customized pre-training.


\subsection{Exploring TokenMoE to Masked AutoEncoder}
\label{sec:tokenmoe}


\paragraph{Overview of TokenMoE.}
TokenMoE~\citep{riquelme2021scaling} is a successful customized supervised
pre-training model
built upon the ViT~\citep{dosovitskiy2020image}, which mainly consists of transformer blocks with alternating multi-head self-attention (MSA) and multi-layer perceptron (MLP).
Specifically, the TokenMoE converts several transformer blocks to Mixture of
Expert (MoE) blocks by expanding the MLP layer $N$ times, each of them is considered as
an {\it expert} (denoted as $E_i(\cdot),\ i=1,2,\dots,N$).
Conditional computation on the $N$ experts is controlled by a {\it gate},
which is a linear layer whose input is the token embedding $\boldsymbol{x}$, and
the output is the top-$K$ probabilities on the experts:
$
G(\boldsymbol{x}) = TopK(\sigma(\boldsymbol{W}_g\boldsymbol{x} + \epsilon))
\label{equ:tokenmoe}
$,
where 
$K$ is the number of experts to be activated,
$\boldsymbol{W}_g$ is the gate parameter, $\sigma$ is the softmax function, and $\epsilon \sim \mathcal{N}(0, \frac{1}{N})$.
$TopK(\cdot)$ returns the $K$ largest entries of
$\sigma(\boldsymbol{W}_g\boldsymbol{x} + \epsilon)$
unchanged but set the others to zero. 
Thus, each token is routed to its corresponding experts. The final output is represented as
\begin{equation}
\bold{y}=\sum_{i=1}^{N} [G(\boldsymbol{x})]_iE_i(\boldsymbol{x}).
\end{equation}
As in~\citep{riquelme2021scaling},
importance loss and load loss are also used to enforce a balanced use of the
experts.
Unless otherwise specified, we set $K=1$ and $N=8$ in all our experiments.


\paragraph{Limitation of TokenMoE.}
As will be shown in the experimental results (Table~\ref{tab:perf_token}), naively
adopting TokenMoE to the MAE cannot
improve performance, even with intense hyper-parameter tuning and 
data augmentations (e.g., Repeat Augment~\citep{Hoffer_2020_CVPR} and
RandAugment~\citep{NEURIPS2020_d85b63ef} with larger magnitude).
Figure~\ref{fig:comp_a}
shows the routing heatmaps of the pre-trained TokenMoE model.
As can be seen, the routing process has  little correlation with the ImageNet labels. 
Moreover, expert 3 
is selected 
most of the time (91.8\% of the classes).
This
degenerates the multi-expert network into a single-expert network.
As demonstrated in Figure~\ref{fig:disadv_tokenmoe}, we speculate that this
is due to the use of low-level pixel values 
(instead of semantic class labels in the original TokenMoE)
as reconstruction
targets.
This is also observed in~\cite{li2022mc}. 



\subsection{Mixture of Cluster-conditional Experts}\label{sec:moce}


To address the limitations of TokenMoE, we propose the Mixture of Cluster-conditional Experts (MoCE), which trains each expert in a semantic-aware manner. 
The procedure consists of data clustering, architecture and gate design, and deployment.

\paragraph{Data Clustering.}
To train each expert semantically, a clustering procedure is first performed to
simulate the label partitioning in Section~\ref{sec:negative_transfer}. 
With a pre-trained MAE model, we collect all the image features 
$\boldsymbol{f_i}$'s
(normalized to unit
length
$\|\boldsymbol{f_i} \|=1$), and
represent
the feature matrix 
as $\boldsymbol{F} = [\boldsymbol{f_1}, \boldsymbol{f_2}, \dots,
\boldsymbol{f_n}] \in \R^{d\times n}$,
where $n$ is the number of images and $d$ is the dimension of the feature.
The learnable cluster centroids are
represented as $\boldsymbol{C}=[\boldsymbol{c_1}, \boldsymbol{c_2}, \dots,
\boldsymbol{c_m}] \in \R^{d\times m}$, (with $\|\boldsymbol{c_i} \|=1$), where
$m$ is the desired number of clusters. 
The assignment of feature to clusters is computed as
$\boldsymbol{A}=\boldsymbol{F}^T\boldsymbol{C}$. Following~\cite{asano2019self},
let $\boldsymbol{Q} \in \mathbb{R}^{m \times n} $ be the posterior distribution
of clustering, whose objective is
\begin{equation}
        \max_{\bm{Q}} \ Tr(\bm{Q}^T\bm{A}) + \epsilon H(\bm{Q}) \quad
        s.t.\quad \bm{Q}\bm{1}_n=\frac{1}{m}\bm{1}_m,\ \bm{Q}^T\bm{1}_m=\frac{1}{n}\bm{1}_n,
    \label{eqn:clustering}
\end{equation}
where $\bm{1}_m$ is the 
$m$-dimensional
vector of all ones,
$H$ is the entropy function, and the constraints force the clustering results to be balanced. 
$\bm{Q}$ and $\bm{C}$ 
are 
optimized 
iteratively. For a given 
$\bm{C}$,
$\bm{Q}$ is solved by the Sinkhorn-Knopp
algorithm~\citep{cuturi2013sinkhorn}; while for a given
$\bm{Q}$,
$\bm{C}$ is obtained by
minimizing the cross entropy between $\bm{Q}$ and $\bm{A}$
with SGD.
We take the final $\bm{C}$ and $\bm{Q}$ as the cluster centroids and clustering
assignments, respectively. The implementation details are in Appendix~\ref{app:clustering}.

\paragraph{Architecture.}
The whole network is trained on the full ImageNet data, with each expert trained
by images from selected clusters decided by the 
MoCE gates'
routing results.
As on average each data cluster has only a fraction of $1/K$ of the original
sample size,
the training time of each expert is also $K$ 
times shorter than the other parameters with dense modeling (e.g.,
MSA parameters~\citep{riquelme2021scaling}), we further adopt a distillation loss
$\mathcal{L}_{distill}$, which is defined as the $\ell_2$ distance between the
features generated by the whole network and each expert. 
This loss function can be formulated as
\begin{equation}
    \min_{\bm{\theta}} \sum_{i=1}^{m} \mathcal{L}_{MAE}(D_i;\bm{\theta_i}) + \mathcal{L}_{distill},
\end{equation}
where 
$D_i$ is the $i$th cluster, $\bm{\theta_i}$ is 
the parameter used for training $D_i$, and
$\mathcal{L}_{MAE}(D_i;\bm{\theta_i})$ is the reconstruction loss for masked image modeling.
$\bm{\theta_i}$ consists of several experts in different layers, as explained in the following.

\paragraph{Gate Design.}

As in the TokenMoE, we replace several MLP layers in the ViT with layers
equipped with MoCE gates. 
In TokenMoE, routings of the tokens to experts 
are considered
separately. In
MoCE,
we route tokens from images of the same cluster to the same expert. 
The MoCE gate output can 
thus
be written as 
\begin{equation}
    G(\textbf{x}) = TopK(\sigma(\bm{W}_g \cdot \bm{C}_{[\textbf{x}]} + \epsilon)),    
    \label{equ:moce}
\end{equation}
where $\bm{W}_g$ is the gate parameter, and
$\bm{C}_{[\textbf{x}]}$ is the embedding of the cluster that $\textbf{x}$ belongs to. 
Empirically, we find that the confidence of $G(\textbf{x})$ (the max entry) is low and consequently, the mapping between clusters and experts varies a lot during pre-training. 
Inspired by the importance and load losses~\citep{riquelme2021scaling}, we
add the following loss $\mathcal{L}_{imbalance}$ to enhance the confidence of the gates. Since it makes $G(\textbf{x})$ shaper, we call it \textit{imbalance} loss.
\begin{equation}
    \mathcal{L}_{imbalance} = - \sum_{i=1}^n\left( \frac{std(G(\textbf{x})_i)}{mean(G(\textbf{x})_i)}\right )^2,
\end{equation}
For practical implementation, the loss is calculated over the samples in a batch. The imbalance loss penalizes on the negative variance of the gate confidence.

\paragraph{Deployment.}
On deployment, customized experts are selected from MoCE, and fine-tuned for each downstream task. 
As shown in Section~\ref{sec:negative_transfer}, we prefer to use the experts that is pre-trained from data whose semantics is closest to that of the downstream task. 
This can be obtained by re-using the data clustering module. 
Specifically, we feed 
images 
for the downstream task
through the pre-trained MAE model and collect all the image features as $\bm{F}_{task}$. 
The assignment of downstream images to the clusters is then computed as
$\bm{A}_{task} = \bm{F}_{task}^T \bm{C}$. We select the largest cluster with
assigned downstream images, and use the corresponding experts 
(a \textbf{sub-model} of the whole MoCE model)
for deployment.
In the case when only one expert is activated
at each MoCE layer ($K=1$), a regular ViT model is needed for downstream fine-tuning, which is much more efficient than MoE.

\section{Experiments}
In this section, we first introduce the setup of pre-training and fine-tuning stage of MoCE in Sec.~\ref{sec:setup}.
Then we demonstrate the effectiveness of MoCE by evaluating the pre-trained models on a collection of 11 downstream tasks with detailed analysis of our MoCE superior to vanilla MAE and TokenMoE in Sec.~\ref{sec:analysis}. 
Finally we take ablation studies on the key components of MoCE in Sec.~\ref{sec:ablation}.

\begin{table}[t!]
  \centering
  \caption{Transfer accuracy (\%) of self-supervised learning models on 11
  downstream tasks.}
  \label{tab:main_table}
  \begin{center}
  \scalebox{0.9}{
  \footnotesize{{
  \setlength{
  \tabcolsep}{0.9mm}
    \begin{tabular}{l|cccccccccccc}
    \toprule
          & \multicolumn{1}{c}{Aircraft} & \multicolumn{1}{c}{Caltech} & \multicolumn{1}{c}{Cars} & \multicolumn{1}{c}{C10} & \multicolumn{1}{c}{C100} & \multicolumn{1}{c}{DTD} & \multicolumn{1}{c}{Flowers} & \multicolumn{1}{c}{Food} & \multicolumn{1}{c}{Pets} & \multicolumn{1}{c}{SUN} & \multicolumn{1}{c}{VOC} & \multicolumn{1}{c}{Avg.} \\
    \midrule
    \multicolumn{13}{l}{{\it ResNet-50}} \\
    \midrule
    BYOL  & 82.39  & 90.12  & 87.33  & 96.28  & 82.15  & 74.57  & 95.96  & 82.13  & 88.52  & 64.41  & 83.97  & 84.35  \\
    DeepCluster-v2 & 78.75  & 90.51  & 86.33  & 96.48  & 82.28  & 75.43  & 96.16  & 83.68  & 90.33  & 66.68  & 81.37  & 84.36  \\
    \midrule
    \multicolumn{13}{l}{{\it Vision Transformer}} \\
    \midrule
    Supervised & 76.55 & 89.98 & 86.19 & 96.79 & 83.96 & 75.09 & 93.94 & 85.17 & 92.54 & 64.54 & 87.22 & 84.72 \\
    DINO  & 66.50  & 91.65  & 76.37  & 98.12  & 86.69  & 75.73  & 96.40  & 93.77  & 93.97  & 59.33  & 86.62  & 84.10  \\
    MoCo v3 & 76.29  & 91.64  & 85.18  & 97.99  & 86.98  & 72.64  & 95.33  & 83.94  & 92.35  & 65.54  & 84.21  & 84.74  \\
    BEiT  & 53.16  & 79.02  & 68.11  & 94.34  & 73.54  & 68.04  & 91.33  & 79.59  & 84.02  & 56.13  & 65.65  & 73.90  \\
    MAE   & 72.38  & 90.47  & 83.51  & 95.69  & 68.40  & 75.48  & 96.10  & 79.98  & 92.35  & 62.43  & 84.79  & 81.96  \\
    \midrule
    MAE*  & 72.71  & 91.24  & 84.47  & 96.15  & 77.33  & 75.05  & 96.25  & 80.49  & 92.78  & 62.46  & 85.02  & 83.09  \\
    MoCE (Ours) & 78.73  & 90.61  & 88.56  & 97.79  & 84.68  & 74.04  & 96.94  & 86.24  & 93.07  & 65.05  & 85.26  & \textbf{85.54}$^{+2.45}$ \\
    \bottomrule
    \end{tabular}%
    }}}%
    \vspace{-0.2cm}
\end{center}
\end{table}%


\begin{table}[t!]
  \centering
  \caption{Transfer accuracy (\%) on detection and segmentation.}
\vspace{0.2cm}
   \scalebox{0.9}{
   \footnotesize{
    \begin{tabular}{l|c|ccc|ccc}
    \toprule
    \multirow{2}[0]{*}{Method} & ADE20K & \multicolumn{6}{c}{COCO} \\
    \cline{2-8}
  & mIoU & AP$^{bb}$ & AP$^{bb}_{50}$ & AP$^{bb}_{75}$ & AP$^{mk}$ & AP$^{mk}_{50}$ & AP$^{mk}_{75}$ \\
    \midrule
    Supervised  & 46.9  & 48.8  & 68.7  & 52.7  & 42.5  & 65.9  & 45.5 \\
    DINO  & 46.9 & 49.5 & 69.1 & 53.6 & 42.9 & 66.0 & 46.3 \\
    MoCo v3 & 46.8 & 47.2 & 66.9 & 50.8 & 41.1 & 63.6 & 44.1  \\
    BEiT & 45.6 & 40.8 & 59.4 & 44.1 & 36.0 & 56.8 & 38.2 \\
    MAE   & 48.1 & 50.6 & 69.4 & 55.0 & 43.8 & 66.6 & 47.5 \\
    \midrule
    MoCE  & \textbf{48.3}  & \textbf{51.1} & \textbf{69.8} & \textbf{55.4} & \textbf{44.2} & \textbf{67.0} & \textbf{48.1} \\
    \bottomrule
    \end{tabular}
  \label{tab:det_seg}}}%
  \vspace{-0.1cm}
\end{table}%


\subsection{Setup}\label{sec:setup}
For all experiments,
we replace two MLP layers with MoCE layers in the original ViT-B~\citep{dosovitskiy2020image}.
Following~\cite{wu2022residual}, layers with the greatest gradient magnitude are
selected (which are the last two MLP layers 
in our
experiments). Unless otherwise specified,
the number of experts is 8
and the number of clusters is 256.
Our model utilizes the officially released 1600-epoch pre-trained MAE
model\footnote{https://github.com/facebookresearch/mae} and continues to train for an extra 200 epochs. 
Each expert is initialized by the corresponding dense model with a small weight
perturbation.
The training procedure mainly follows that of MAE, except that we multiply the
base learning rate by 0.1. All regularization loss
weight is set to 0.01 by default. 

To 
ensure a fair comparison with the vision transformer
on downstream classification tasks,
we mainly follow the hyper-parameter settings in 
~\citep{dosovitskiy2020image,riquelme2021scaling} and the benchmark settings
in~\citep{ericsson2021well}. 
The proposed
model is
compared
with various
self-supervised models, including DINO~\citep{caron2021emerging}, MoCo v3~\citep{chen2021empirical},
BEiT~\citep{bao2021beit}, and the highly-performant ResNet-50 models of
BYOL~\citep{byol20} and DeepCluster-v2~\citep{caron2018deep}.  We also compare with the supervised pre-trained model DeiT~\citep{pmlr-v139-touvron21a}. 
To make a fair comparison of training time, we continue to train a 1600-epoch pre-trained MAE for 200 epochs with total ImageNet as our baseline, and is denoted as MAE* in Table~\ref{tab:main_table}.
For detection and segmentation tasks, 
following \cite{bao2021beit},
we perform experiments on ADE20K~\citep{ADE20K} and COCO~\citep{lin2014microsoft}.
We utilize the officially released checkpoints for all baselines. 
Details are in
Appendix~\ref{app:eva_detail}.

\subsection{Results}\label{sec:analysis}

\paragraph{Transfer Results.} The classification transfer performance of various
self-supervised models are shown in Table~\ref{tab:main_table}. As can be seen, MoCE achieves a 2.45\% improvement over MAE*
and reaches the
state-of-the-art averaged accuracy, demonstrating the effectiveness of the
task-customized pre-training paradigm. 
On fine-grained datasets such as {\it
Aircraft}, {\it Cars} and {\it Food}, MoCE outperforms the baseline model by a large margin. 
This is because these fine-grained tasks are similar
to only a subset of the pre-training dataset. Hence, MoCE 
can alleviate negative transfer by
using
the model that is trained by the cluster most similar to the particular downstream
task.
On the other hand,
MoCE shows only limited improvement
on tasks such as {\it Caltech}, {\it Cifar-10} and {\it VOC}.
These tasks are more general and 
contain images covering the various semantics in the pre-training dataset, and
thus negative
transfer does not exist.

Table ~\ref{tab:det_seg} shows the transfer performance on  the
detection and segmentation tasks. As can be seen,
MoCE outperforms MAE 
  and the other  baselines (including the supervised one),
and achieves state-of-the-art results.


\begin{table}[tbp]
    \centering
    \caption{Transfer accuracy of MAE, TokenMoE, SDR and MoCE. 
    SDR(ViT) is our re-implementation of SDR under ViT.
    We observe that TokenMoE cannot outperform vanilla MAE, while SDR(ViT) achieves better performance, which is further outperformed by MoCE.
    }
    \begin{center}
    \scalebox{1}{
    \footnotesize{{
    \setlength{\tabcolsep}{1mm}
    \begin{tabular}{l|rrrrrrrrrrrr}
        \toprule
        & \multicolumn{1}{c}{Aircraft} & \multicolumn{1}{c}{Caltech} & \multicolumn{1}{c}{Cars} & \multicolumn{1}{c}{C10} & \multicolumn{1}{c}{C100} & \multicolumn{1}{c}{DTD} & \multicolumn{1}{c}{Flowers} & \multicolumn{1}{c}{Food} & \multicolumn{1}{c}{Pets} & \multicolumn{1}{c}{SUN} & \multicolumn{1}{c}{VOC} & \multicolumn{1}{c}{Avg.} \\
        \midrule
        MAE*   &  72.71  & 91.24  & 84.47  & 96.15  & 77.33  & 75.05  & 96.25  & 80.49  & 92.78  & 62.46  & 85.02  & 83.09  \\
        TokenMoE & 70.51  & 89.70  & 81.40  & 95.18  & 76.44  & 73.67  & 95.09  & 77.45  & 90.71  & 61.12  & 80.15  & 81.04  \\
        \midrule
        {SDR} & {75.77} & {89.73} & {86.65} & {95.31} & {83.60}  & {73.62} & {95.53} & {84.77} & {91.25} & {64.64} & {83.51} & {84.03} \\
        {SDR(ViT)} &  76.57  & 90.04  & 86.95  & 96.92  & 81.42  & 73.09  & 96.14  & 82.90  & 92.65  & 64.40  & 85.37  & 84.22  \\
        MoCE & 78.73  & 90.61  & 88.56  & 97.79  & 84.68  & 74.04  & 96.94  & 86.24  & 93.07  & 65.05  & 85.26  & \textbf{85.54} \\
        \bottomrule
    \end{tabular}}}}%
    \label{tab:perf_token}%
\vspace{-0.7cm}
\end{center}
\end{table}%


\paragraph{Comparison between MoCE, TokenMoE, MAE and SDR.} 
In this experiment, we compare MoCE  with the following
models:
(i)
MAE, (ii) TokenMoE, (iii)
SDR~\citep{liu2022task}, a task-customized model
that aims at alleviating negative transfer, and (iv) 
SDR(ViT), which
re-implements
SDR
with the ViT architecture.
Table~\ref{tab:perf_token} shows the transfer accuracy
on 11 downstream tasks.
As can be seen, TokenMoE performs even worse than MAE, suggesting that naively
adopting MoE to MAE is not desirable. 
Both MoCE and SDR(ViT) outperform MAE, demonstrating the effectiveness of task-customized methods for alleviating negative transfer. 
MoCE further outperforms SDR(ViT), indicating the importance of self-adaptive routing.

\begin{figure}[!tbp] 
\centering
\subfigure[Routing heatmap for TokenMoE experts.]{
 \begin{minipage}[b]{0.5\linewidth}
    \centering
    \includegraphics[width = 0.6\linewidth]{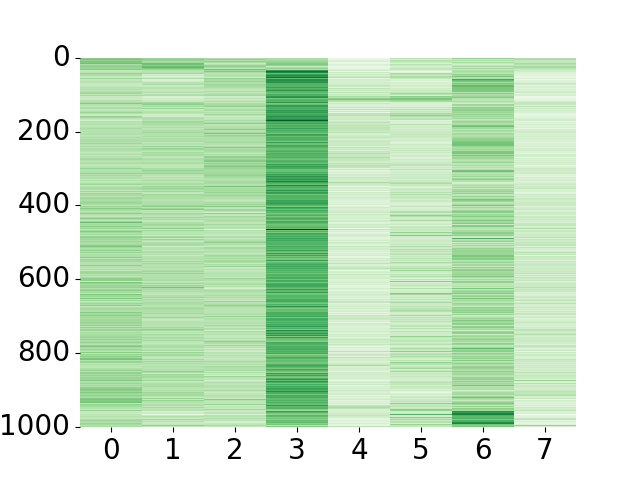}
     \label{fig:comp_a}
\end{minipage}
}\subfigure[Examples pre-training samples for expert 1 (top), expert 2 (middle), and
expert 3 (bottom).]{
 \begin{minipage}[b]{0.5\linewidth}
    \centering
    \includegraphics[width = 0.7\linewidth]{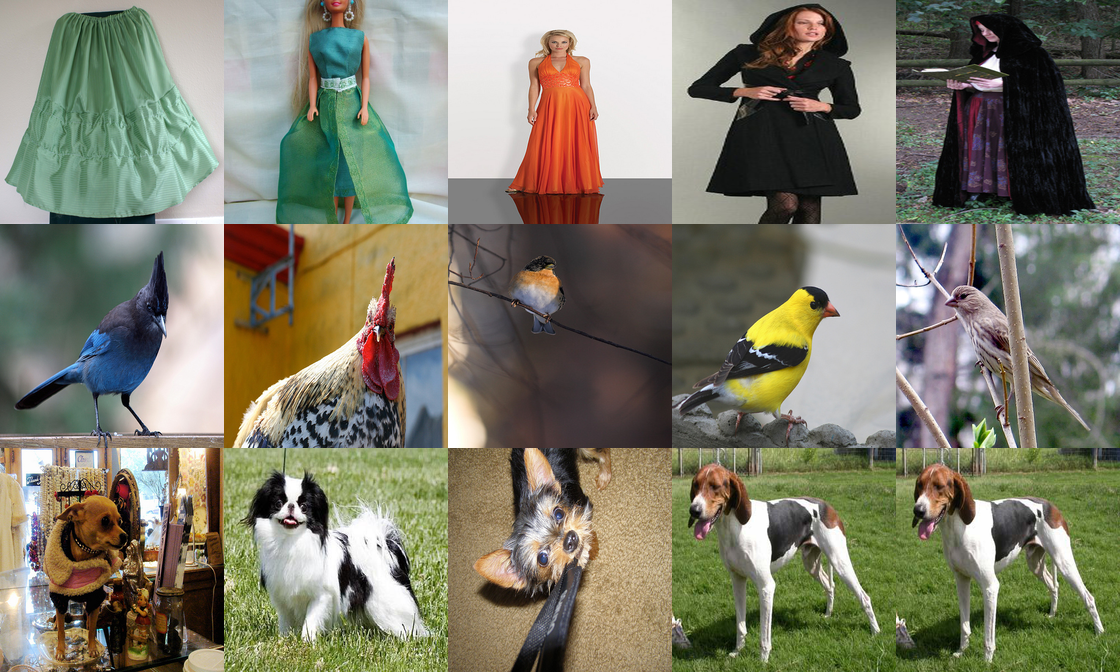}
    \label{fig:sample_image}
    \end{minipage}}

\subfigure[Routing heatmap for MoCE experts.]{
 \begin{minipage}[b]{0.5\linewidth}
    \centering
    \includegraphics[width = 0.6\linewidth]{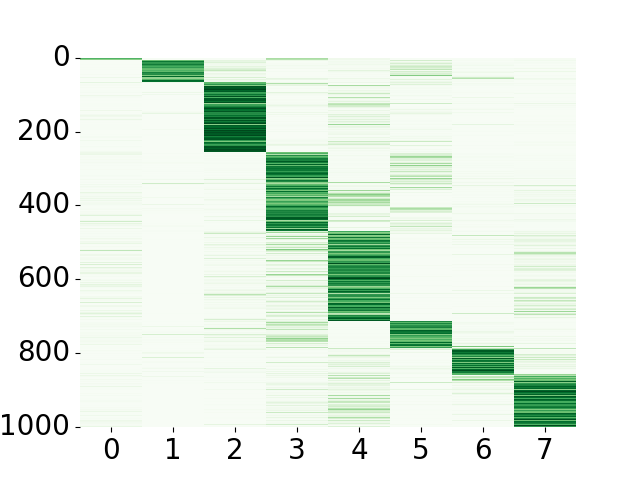}
    \label{fig:comp_b}
\end{minipage}
}\subfigure[Relative PSNR improvement over MAE.]{
 \begin{minipage}[b]{0.5\linewidth}
    \centering
    \includegraphics[width = 0.8\linewidth]{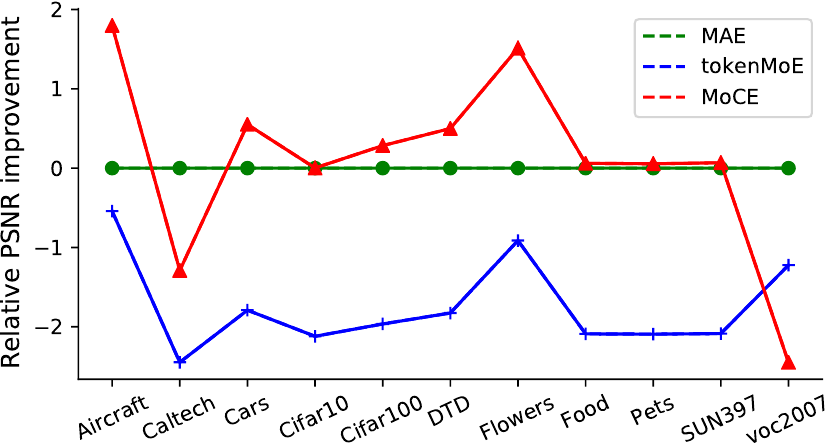}
    \label{fig:psnr}
\end{minipage}}
\caption{
(a),(c): Routing heatmaps for experts in TokenMoE and MoCE. 
The x-axis is the expert ID, and the y-axis is the ImageNet semantic label ID. 
Darker green means a higher proportion of tokens belonging to the corresponding
class are allocated to the expert. 
The label is sorted differently in each figure to make it readable.
(b): Example samples from the pre-training dataset of 3 MoCE experts. 
(d): Relative PSNR improvement of
TokenMoE and MoCE 
over MAE
for each downstream task. 
}
\end{figure}

Figure~\ref{fig:psnr} 
shows the peak signal-to-noise ratio
(PSNR)~\citep{sara2019image}, which reflects the generation quality of these autoencoder models.
MoCE exhibits improvement over TokenMoE and MAE on most datasets.
We also provide the comparisons in the case of a fair parameter count, large
architectures, and training from scratch in the Appendix~\ref{app:param},~\ref{app:large} and~\ref{app:scratch}, respectively.



\paragraph{Analysis on experts.}
Figure~\ref{fig:comp_a} and
Figure~\ref{fig:comp_b} 
show the routing heatmaps for
TokenMoE and
MoCE, respectively. As can be seen,
routing of
the TokenMoE experts
has little correlation with semantics. On the other hand,
each MoCE expert is
trained by several classes, showing a more balanced assignment of images to experts. 
This verifies that the improvement of MoCE is due to more effective learning of
the experts. 
Moreover,
notice that 
the importance loss and load balance loss~\citep{riquelme2021scaling} are applied
and indeed work as “expected” because they are only applied with respect to patch tokens instead of semantic classes.
On the other hand, MoCE can balance the experts both at the token level and semantic level.

Figure~\ref{fig:sample_image}
shows example pre-training samples for 3 random MoCE experts.
Note that expert 1 is mostly
trained by images containing clothes, experts 2 
is pre-trained mostly by
bird images, while
expert 3 is pre-trained mostly by
dog images.

Next, we show that each expert is
trained by samples with similar semantics. Following~\citep{mikolov2013efficient}, we select
the label set used by each expert, and
then 
compute the $\ell_2$ distances between 
CLIP embeddings \citep{2021Learning} 
of 
labels
used by the same expert and by different experts.
The average distance between labels used by the same expert is 0.84, while that
between labels used by different experts is 
0.92,
indicating that the MoCE gate automatically aggregates labels with similar
semantics to each expert, thus benefiting downstream transfer.

\paragraph{Training and testing efficiency.}

Table~\ref{tab:efficiency}
compares the
efficiencies of MAE,
TokenMoE and MoCE
during training and testing.
As can be
seen,
all
of them have similar FLOPs. However, TokenMoE needs to use the whole
model during both training and testing, while MoCE only needs to use
a single expert
and thus halves the required number of parameters when
testing. In addition, the training and testing speeds are improved by respectively 18\% and 37\%, which is attributed to the reduction of token shuffle operations as
tokens in the same image do not need to be split and are dispatched to the same expert,
significantly reducing the communication overhead.


\subsection{Ablation}\label{sec:ablation}

\paragraph{Search method.}
When a downstream task arrives, it is expensive to fine-tune all experts to choose the best one.
To find the task-customized expert ($K=1$),
we compare 
the method proposed in Section~\ref{sec:moce} with
(i) early stop, (ii)
KNN~\citep{liu2022task}, (iii) LogME~\citep{you2021logme}.
The experiment is performed on the task with the most samples (\textit{Food}),
the task with  the least samples (\textit{Flowers}), and the one with a medium number of samples (\textit{Aircraft}).
For comparison,
we additionally show the performance of the best and worst experts based on an exhaustive search.\
As can be seen in Table~\ref{tab:search},
MoCE performs stably
among different sizes of the dataset, and the search cost is negligible as we only
need to infer the downstream task once and feed it to the clustering module. This
illustrates another advantage of
combining clustering and pre-training in a single paradigm.

\paragraph{MoCE Architectures.} In this experiment, we study the different architecture hyper-parameters of MoCE in three aspects. 
First, we vary the number of experts in each MoCE layer. As can be seen from
Table~\ref{tab:experts}, using more experts leads to consistent improvement on the
accuracy.

Next, we vary the location of the MoCE layers. 
As mentioned in Section~\ref{sec:setup}, we select the MoCE layers based on the
gradient magnitudes. 
In the experiments, MoCE 
selects 
the 11th and 12th MLP layers.
On the other hand, TokenMoE chooses
the last 2 even-numbered 
(i.e., 10th and 12th)
MLP layers.
Furthermore, we exhibit the performance with only 1 and 4 MoCE layers, which are also selected based on the gradient magnitudes.
As shown in Table~\ref{tab:layers}, 
we notice that it is essential to choose the right layer to be MoCE layer. Adding more MoCE layers shows little improvement. 

We also train MoCE with different numbers of clusters. 
As shown in Table~\ref{tab:clusters}, the accuracy increases up to 
256 clusters, and then begins to drop. We hypothesize that with a moderate number of clusters, MoCE can produce a variety of task-customized models.
With even more clusters, the number of experts become the bottleneck and performance starts to saturate.  


\begin{table}[!tb]
    \begin{minipage}{.43\textwidth}
\caption{Efficiency during training (top) and testing (bottom).}
        \vspace{1mm}
        \centering
        \setlength{\tabcolsep}{1.0mm}
        \begin{tabular}{c|c|cc}
            \toprule
            & {MAE} & TokenMoE & MoCE \\
            \midrule
            Params (M) & {111.91} & 178.03 & 178.03 \\
             FLOPs (G) & {9.80} & 9.81 & 9.81 \\
            Speed$\uparrow$ & {1.41x} & 1x & \textbf{1.18x} \\
            \midrule
            \# Params (M) &  {85.88} & 152.00 & 85.88 \\
             FLOPs (G) & {16.88} & 16.88 & 16.88 \\
            Speed$\uparrow$ & {1.37x} & 1x & \textbf{1.37x} \\
            \bottomrule
            \end{tabular}%
            \label{tab:efficiency}%
    \end{minipage}%
    \hspace{2.5mm}
    \begin{minipage}{.53\textwidth}
        \centering
        \caption{The
		  search cost (in GPU hours)
		  for different expert search algorithms.}
        \vspace{1.7mm}
        \setlength{\tabcolsep}{0.6mm}
        \begin{tabular}{l|cccc}
        \toprule
        & \multicolumn{1}{l}{Aircraft} & \multicolumn{1}{l}{Flowers} & \multicolumn{1}{l}{Food} & GPU hours \\
        \midrule
        Best  & 79.92 & 97.96 & 86.24 & 288 \\
        Worst & 69.84 & 94.97 & 81.51 & 288 \\
        \midrule
        Early stop & 77.00  & 96.83 & 85.33 & 144 \\
        KNN   & 71.40  & 95.10  & 83.32 & 5 \\
        LogME & 73.84 & 96.54 & 85.11 & 5 \\
        MoCE  & \textbf{78.73} & \textbf{96.94} & \textbf{86.24} & \textbf{1} \\
        \bottomrule
        \end{tabular}%
    \label{tab:search}%
    \end{minipage} 
\end{table}


\begin{table}[!tb]
\begin{minipage}[t]{0.3\textwidth}
\makeatletter\def\@captype{table}
\caption{Accuracies with different numbers of experts in a MoCE layer.
(default setting used is in bold).}
\centering
\begin{tabular}{c|c}
\toprule
    \# experts & Acc (\%) \\
    \midrule
    1     & 83.09 \\
    2     & 83.01 \\
    4     & 84.22 \\
    \textbf{8} & \textbf{85.54} \\
    \bottomrule
    \end{tabular}%
\label{tab:experts}
\end{minipage}
\hspace{1.8mm}
\begin{minipage}[t]{0.33\textwidth}
\makeatletter\def\@captype{table}
\caption{Accuracies with different numbers and locations of the MoCE layers
(default setting used is in bold).}
\centering
\begin{tabular}{c|c}
\toprule
    \# MoCE layers & Acc (\%) \\
    \midrule
    1     & 83.09 \\
    2 (10th \& 12th) & 83.08 \\
    \textbf{2 (11th \& 12th)} & \textbf{85.54} \\
    4     & 85.59 \\
    \bottomrule
    \end{tabular}%
\label{tab:layers}
\end{minipage}
\hspace{1.8mm}
\begin{minipage}[t]{0.3\textwidth}
\makeatletter\def\@captype{table}
\caption{Accuracies with different numbers of clusters
(default setting used is in bold).}
\vspace{1mm}
\centering
\begin{tabular}{c|c}
\toprule
    \# clusters & Acc (\%) \\
    \midrule
    16    & 82.00 \\
    64    & 84.02 \\
    \textbf{256} & \textbf{85.54} \\
    512   & 85.33 \\
    \bottomrule
    \end{tabular}%
\label{tab:clusters}
\end{minipage}
\end{table}


\section{Conclusion}
In this work, we first show that the negative transfer phenomenon exists in the prevailing self-supervised learning method MAE through extensive experiments. It will impede the scalability of MAE as more pre-training data may instead degenerate the downstream performance. In order to tackle the problem, we introduce Mixture of Expert to MAE as the multi-experts design can equip MAE with different ability that aids transfer. However, different from supervised pre-training, TokenMoE suffers from the fact that the gate shows no correlation to the semantics and the transfer ability is not improved. Based on this, we propose MoCE to explicitly train each expert with different clusters through the MoCE gate design and several losses to stabilize the training process.
A search algorithm for selecting the best model for transfer is also proposed based on the clustering priors. Extensive experiments show that MoCE trains each expert with meaningful semantics and achieves state-of-the-art transfer performance on a collection of 11 downstream tasks and both detection and segmentation tasks.
It is the first work that successfully trains a self-supervised learning MoE model on ImageNet only. We hope such a design will motivate more research on the self-supervised MoE models.

\subsubsection*{Acknowledgments}
We gratefully acknowledge the support of MindSpore, CANN (Compute Architecture for Neural Networks) and Ascend AI Processor used for this research.

\newpage

\bibliography{iclr2023_conference}
\bibliographystyle{iclr2023_conference}

\newpage
\appendix
\section{Appendix}
\subsection{Details for clustering.}
\label{app:clustering}
For data clustering, the features are computed by inferring the pre-train MAE, and
the matrix
$\textbf{Q}$ and $\textbf{C}$ are solved by the Sinkhorn-Knopp algorithm and SGD optimizer iteratively. For the Sinkhorn-Knopp algorithm, we set the iteration number as 3. The learning rate of SGD is set to 0.1, the momentum is 0.9 and weight decay is set to 0.9 for the sparse assignment of cluster results. We train 10 epochs in total and it costs 3 minutes and 20 seconds on average for a single GPU.

\subsection{Comparison under fair parameter counts.}
\label{app:param}
The setting used in our work focuses on a fair comparison of \textbf{FLOPs}, referring to Table~\ref{tab:efficiency} in the main paper.
Since TokenMoE and MoCE always activate only one expert throughout the whole pre-training and fine-tuning procedure, the FLOPs value is maintained close to MAE. 
Apart from this criterion, we further provide the comparison on equal \textbf{parameter} counts. As shown in Table~\ref{tab:equal_param}, we train MAE under the same parameter count as the whole model of MoCE, and MoCE still outperforms MAE consistently.

\begin{table}[!ht]
  \centering
  \caption{Comparison of MAE and MoCE under equal parameter counts. We train MAE with a larger model that shares the same parameter count as the whole model of MoCE.}
    \setlength{\tabcolsep}{1mm}
    \resizebox{\textwidth}{!}{
    \begin{tabular}{l|c|cccccccccccc}
    \toprule
     & \# Params & Aircraft & Caltech & Cars & C10 & C100 & DTD & Flowers & Food & Pets & SUN & VOC & Avg. \\
     \midrule
    MAE & 178.03 & 74.43 & 90.30  & 85.50  & 96.90  & 83.80  & 74.84 & 96.30  & 81.86 & 92.97 & 62.98 & 85.51 & 84.13 \\
    MoCE & 178.03 & 78.73  & 90.61  & 88.56  & 97.79  & 84.68  & 74.04  & 96.94  & 86.24  & 93.07  & 65.05  & 85.26  & \textbf{85.54} \\
    \bottomrule
    \end{tabular}}%
  \label{tab:equal_param}%
\end{table}%

\subsection{MoCE for larger architecture.}
\label{app:large}
{Here we provide the analysis on the larger architecture(ViT-L, 2.57× larger than ViT-B) to explore the scalability of MoCE. We first demonstrate that negative transfer still exists for larger architecture by training MAE with total ImageNet, Split-A and Split-B following the same setting in Sec.~\ref{sec:negative_transfer}. As shown in the first three rows of Table.~\ref{tab:large_arch}, a similar phenomenon is observed that MAE-L trained by Split-A performs better in \textit{Aircraft}, \textit{Cars}, \textit{DTD} and \textit{SUN} while Split-B in \textit{Flowers}, \textit{Food}, and \textit{Pets}. On the other hand, MoCE-L can still alleviate the problem and therefore transfers better. We believe that the negative transfer phenomenon mainly exists when a common pre-trained model is used for various downstream tasks, due to the inevitable semantic gaps between the pre-training and downstream datasets, rather than the architecture. }

\begin{table}[!ht]
  \centering
  \caption{Comparison of MAE and MoCE on ViT-L. We also train MAE with 2 subsets of ImageNet, namely Split-A and Split-B, following the same setting mentioned in Sec.~\ref{sec:negative_transfer}. This table shows that negative transfer still exists on larger architectures, while MoCE can alleviate this problem and achieve better transfer results.}
\resizebox{\textwidth}{!}{
    \setlength{\tabcolsep}{1mm}
    \begin{tabular}{l|cccccccccccc}
    \toprule
     & Aircraft & Caltech & Cars & C10 & C100 & DTD & Flowers & Food & Pets & SUN & VOC & Avg. \\
     \midrule
    MAE-L (full set) & 74.30 & 93.97 & 88.60 & 97.85 & 82.47 & 77.61 & 96.67 & 81.22 & 93.97 & 67.99 & 88.30 & 85.72\\
    MAE-L (Split-A) & 79.70 & 91.59 & 89.33 & 96.97 & 80.38 & 78.67 & 95.44 & 82.97 & 92.49 & 68.73 & 82.41 & 85.33 \\
    MAE-L (Split-B) & 73.42 & 90.80 & 86.00  &  96.18 & 78.73 & 77.34 & 96.75 & 83.63 & 94.92 & 66.06 & 85.85 & 84.52 \\
    MoCE-L & 87.04 & 94.86 & 90.72 & 98.29 & 87.49 & 76.65 & 97.38 & 88.21 & 95.89 & 69.49 & 89.13 & \textbf{88.65}$^{+2.93}$  \\
    \bottomrule
    \end{tabular}}%
  \label{tab:large_arch}%
\end{table}%

\subsection{Performance of MoCE without pre-training.}
\label{app:scratch}
We provide results of MoCE trained from scratch for 200 epochs and 1600 epochs in Table~\ref{tab:scratch}. In this experiment, for clustering, we first pre-train MAE for 50 epochs and perform clustering. We then train MoCE from scratch for 200 epochs and 1600 epochs based on the clustering results.
Although it is a common practice to utilize pre-trained dense models as initialization to accelerate pre-training~\citep{wu2022residual,bai2022masked}, MoCE still outperforms MAE consistently in various downstream tasks when trained from scratch.

\begin{table}[!ht]
  \centering
  \caption{Comparison of MAE and MoCE both training from scratch for 200 epochs (first two rows) and 1600 epochs (last two rows).}
 \scalebox{1}{
    \footnotesize{{
    \setlength{
    \tabcolsep}{0.85mm}
    \begin{tabular}{l|cccccccccccc}
    \toprule
     & Aircraft & Caltech & Cars & C10 & C100 & DTD & Flowers & Food & Pets & SUN & VOC & Avg. \\
     \midrule
    MAE   & 64.73 & 85.91 & 77.10  & 92.92 & 72.50  & 73.30  & 93.11 & 73.14 & 88.70  & 57.84 & 73.27 & 77.50 \\
    MoCE  & 71.16 & 90.55 & 82.46 & 96.06 & 76.56 & 74.57 & 95.70  & 79.67 & 92.58 & 62.20  & 84.25 & \textbf{82.34} \\
    \midrule
    MAE & 72.38  & 90.47  & 83.51  & 95.69  & 68.40  & 75.48  & 96.10  & 79.98  & 92.35  & 62.43  & 84.79  & 81.96  \\
    MoCE & 78.75  & 91.64  & 87.04  & 97.15  & 83.12  & 73.62  & 96.08  & 83.84  & 93.06  & 65.49  & 85.81  & \textbf{85.05}  \\
    \bottomrule
    \end{tabular}}}}%
  \label{tab:scratch}%
\end{table}%

\subsection{Evaluation details for downstream tasks.}
\label{app:eva_detail}
\paragraph{Classification.} We mainly follow the settings of~\cite{ericsson2021well}.to make a fair comparison. Specifically, all models are trained by SGD with a momentum of 0.9. Weight decay is set to be 0 and the learning rate is searched among [1e-4, 3e-4, 1e-3, 3e-3, 1e-2, 3e-2, 1e-1, 3e-1]. 
Each model is fine-tuned for 2500 steps with cosine learning rate decay, a batch size of 64, and 224$\times$224 resolution.
We fine-tune each model 3 times and report the average performance. We find such a setting generates a stable result. 

\paragraph{Semantic segmentation.}
We evaluate MoCE on the semantic segmentation task that aims to predict the class for each pixel in the image. We report the metric of mean Intersection of Union (mIoU) averaged over all semantic categories in ADE20K~\citep{ADE20K}. We choose the best expert by applying ADE20K images to our clustering module and selecting the cluster that contains the most images. We use Adam~\citep{loshchilov2017decoupled} as the optimizer. The learning rate is set to 1e-3 with layer-wise learning rate decay~\citep{clark2020electra} to be 0.65. We conduct fine-tuning for 160K steps. The batch size is 16. The detailed hyper-parameters can refer to \cite{bao2021beit}. 

\paragraph{Detection and Instance segmentation } are also evaluated on COCO~\citep{lin2014microsoft}. We follow the same deployment method as the one used in the semantic segmentation task to choose the best expert. Following iBOT~\citep{ibot} we adopt the Cascade Mask R-CNN~\citep{cai18cascadercnn,he2017mask} and the multi-scale training. The shorter side is randomly resized between 480 and 800 while the longer one is no longer than 1333. The batch size is 16, and the initial learning rate is 1e-4. The layer-wise learning rate decay ratio~\citep{clark2020electra} is set to 0.75. We train the model for 12 epochs and decrease the learning rate by 10x at epoch 9 and 11.

\end{document}